\pdfoutput=1

\documentclass[11pt]{article}

\usepackage[]{ACL2023}

\usepackage{times}
\usepackage{latexsym}

\usepackage[T1]{fontenc}

\usepackage[utf8]{inputenc}

\usepackage{microtype}

\usepackage{inconsolata}

\usepackage{amsmath} 
\usepackage[capitalise,noabbrev]{cleveref}
\usepackage{graphicx} 
\usepackage[normalem]{ulem} 

\newcommand\blfootnote[1]{%
  \begingroup
  \renewcommand\thefootnote{}\footnote{#1}%
  \addtocounter{footnote}{-1}%
  \endgroup
}

\title{Long-form Simultaneous Speech Translation$^*$ \\ \large Thesis Proposal}

\author{Peter Pol\'ak \\
  Charles University \\
  Faculty of Mathematics and Physics \\
  Institute of Formal and Applied Linguistics \\
  \texttt{polak@ufal.mff.cuni.cz} \\}

\begin{document}
\maketitle
\begin{abstract}
Simultaneous speech translation (SST) aims to provide real-time translation of spoken language, even before the speaker finishes their sentence. Traditionally, SST has been addressed primarily by cascaded systems that decompose the task into subtasks, including speech recognition, segmentation, and machine translation. However, the advent of deep learning has sparked significant interest in end-to-end (E2E) systems. Nevertheless, a major limitation of most approaches to E2E SST reported in the current literature is that they assume that the source speech is pre-segmented into sentences, which is a significant obstacle for practical, real-world applications. This thesis proposal addresses end-to-end simultaneous speech translation, particularly in the long-form setting, i.e., without pre-segmentation. We present a survey of the latest advancements in E2E SST, assess the primary obstacles in SST and its relevance to long-form scenarios, and suggest approaches to tackle these challenges. 
\end{abstract}

\blfootnote{$^*$The literature on simultaneous speech translation often uses the word ``streaming'' as an equivalent of ``simultaneous'' to refer to the translation of an unfinished utterance. In other literature, however,  the term ``streaming'' refers to input spanning several sentences. To avoid confusion, we use ``simultaneous'' to refer to the translation of an unfinished utterance and ``long-form'' to refer to input spanning several sentences.}

\vspace{-2.5em}
\section{Introduction}
In today's highly globalized world, communication among individuals speaking different languages is gaining importance. International conferences and multinational organizations like the European Parliament often rely on human interpreters. However, in many scenarios, employing human interpreters can be impractical and costly. In such cases, simultaneous speech translation\footnote{We consider only the speech-to-text variant in this work.} (SST) offers a viable solution by enabling real-time translation before the speaker completes their sentence. 

Traditionally, both offline speech translation (ST) and simultaneous speech translation (SST) have relied predominantly on cascaded systems that decompose the task into multiple subtasks, including speech recognition, speech segmentation, and machine translation \cite{225935, fugen2007simultaneous, bojar-etal-2021-elitr}. However, recent advancements in deep learning and the availability of abundant data \cite{tan2018artificial, sperber-paulik-2020-speech} have led to a significant paradigm shift towards end-to-end (E2E) models. While the cascaded approach continues to dominate offline ST, the opposite is true for SST \cite{anastasopoulos-etal-2022-findings, agrawal-etal-2023-findings}.

Despite the recent popularity of end-to-end SST, the vast majority of research focuses on the ``short-form'' setting, which assumes that the speech input is already pre-segmented into sentences. Critically, this assumption poses an obstacle to deployment in the wild. Therefore, we aim to achieve a ``true'' long-form simultaneous speech translation in our thesis. We break down our efforts into three steps:

\paragraph{Quality-latency tradeoff in SST}
The first step of our research concentrates on enhancing the quality-latency tradeoff, mainly in the traditional ``short-form'' regime. We will evaluate different approaches and architectures.

\paragraph{Towards the long-form SST}
In the next step, we will explore the feasibility of long-form simultaneous speech translation by adopting segmented inference. 

\paragraph{True long-form SST}
The final goal of our work is to explore the potential of end-to-end modeling for true long-form SST. We will focus on identifying an appropriate model architecture and effective training procedures to achieve seamless and reliable long-form simultaneous speech translation.

The next section introduces some important aspects of simultaneous speech translation.

\section{Simultaneous Speech Translation}
\label{sec:sst}

The ultimate goal of SST is to enable \emph{real-time} communication between people speaking different languages. To achieve this goal, SST systems must meet two important criteria. First, they must be \emph{computationally efficient} to ensure timely translation during ongoing speech. 
Second, SST systems must be capable of \emph{handling unfinished sentences}. Working with unfinished sentences allows for more timely translations, particularly when waiting for sentences to be completed is impractical, such as matching slides or presenters' gestures. However, translating unfinished sentences increases the risk of translation errors since translation usually requires re-ordering that benefits from a more complete sentence context. Thus, there exists a \emph{quality-latency tradeoff}. This means that given a certain latency constraint, we want the model to produce as good translations as possible. Ideally, we want the model to ``predict'' the future context without the risk of an incorrect translation. The quality-latency tradeoff is one of the main topics of our research. %

\subsection{Re-Translation vs. Incremental SST}

SST can be classified as either re-translation or incremental. Re-translation SST \cite{NiehuesNguyenCho2016_1000062876, NiehuesPhamHa2018_1000087584} can revise the hypothesis or re-rank the set of hypotheses as more speech input is read. Revising the translation allows the re-translation SST to have comparable final translation quality with the offline speech translation \cite{arivazhagan-etal-2020-translation}. This design approach arguably introduces challenges for the user in processing the translation and makes it impossible to use in real-time speech-to-speech translation. Additionally, it also complicates the latency evaluation.

In fact, several SST latency metrics \cite{ma-etal-2020-simuleval} were originally developed specifically for incremental translation scenarios.\footnote{IWSLT shared tasks \cite{ansari-etal-2020-findings, anastasopoulos-etal-2021-findings, anastasopoulos-etal-2022-findings} also follow this evaluation standard.}
Incremental SST \cite{cho2016can,dalvi-etal-2018-incremental} differs from the re-translation system in that it prunes all hypotheses to a common prefix, which is then shown to the user. For the user, the translation changes only by incrementally getting longer; none of the previously displayed outputs are ever modified. In our work, we focus on incremental SST.

\subsection{Cascaded vs. End-to-End}
\label{sec:cascade-vs-e2e}
Traditionally, offline speech translation and SST were achieved as a \emph{cascade} of multiple systems: automatic speech recognition (ASR), inverse transcript normalization, which includes punctuation prediction and true casing, and machine translation (MT, \citealp{225935, fugen2007simultaneous, bojar-etal-2021-elitr}). The advantage of the cascade approach is that we can optimize models for each subtask independently. Also, ASR and MT tasks typically have access to larger and more diverse corpora than direct speech translation.

However, using a cascade system introduces several challenges \cite{sperber-paulik-2020-speech}. The most important among them is \emph{error propagation} \cite{ruiz2014assessing}. Further, MT models might suffer from \emph{mismatched domains} when trained on written language. Furthermore, as the source is transformed into a textual form, it \emph{loses crucial information about prosody}, i.e., the rhythm, intonation, and emphasis in speech \cite{bentivogli-etal-2021-cascade}. %
Finally, many languages, especially endangered ones, have no written form, which makes the cascade approach impractical or impossible for such languages \cite{harrison2007languages,duong-etal-2016-attentional}.

As of the latest findings, the current state-of-the-art for offline speech translation continues to be based on a cascaded approach \cite{anastasopoulos-etal-2022-findings,agrawal-etal-2023-findings}. In simultaneous speech translation, however, both approaches yield competitive performance. %
The advantage of the end-to-end models in SST may be that they avoid the extra delay caused by ASR-MT collaboration in the cascade \cite{wang-etal-2022-hw-tscs}. 

In our work, we focus on end-to-end models.

\section{Long-form Simultaneous Speech Translation}
\label{sec:longform}

Most of the contemporary research on SST assumes speech pre-segmented into short utterances with segmentation following the sentence boundaries. However, in any real application, there is no such segmentation available. This section places long-form SST within the broader context of long-form ASR, MT, and offline ST. Subsequently, we explore the current literature on long-form SST.

\subsection{Long-Form ASR}
In terms of input and output modalities, long-form ASR and ST face similar issues. There are two types of strategies for long-form processing: (1) the \emph{segmented approach}, which divides the input into smaller chunks, and (2) the \emph{true long-form approach}, which handles the entire long-form input as a single unit.

Most of the literature focuses on the \emph{segmented approach}. A typical solution involves pre-segmenting the audio using voice activity detection (VAD). However, VAD segmentation may not be optimal for real-world speech since it might fail to handle hesitations or pauses in sentences that must be treated as undivided units. More sophisticated approaches leverage latent alignments obtained from CTC \cite{graves2006connectionist} and RNN-T \cite{graves2012sequence} for better segmentation \cite{yoshimura2020end, huang2022e2e}. Alternatively, segmentation into \emph{fixed segments} is also popular \cite{chiu2019comparison, chiu2021rnn}. To reduce low-quality transcripts close to the segment boundaries, they typically perform overlapped inference and use latent alignments to merge the transcripts correctly. The chunking approach is also adopted by the attentional model Whisper in the offline \cite{radford2023robust} and simultaneous regime \cite{machacek-etal-2023-turning}. 

Another line of work focused on \emph{long-form modeling} directly. For example, \citet{chiu2019comparison} conducted a comprehensive study comparing different architectures, including RNN-T and attention-based models. The findings indicate that only RNN-T and CTC architectures can generalize to unseen lengths. To further improve the true long-form ASR, \citet{narayanan2019recognizing} suggest simulation of long-form training by LSTM state passing.

While the previously mentioned research was predominantly based on RNNs, more recent work has transitioned to utilizing Transformer models. \citet{zhang2023google} compared a chunk-wise attention encoder, which involves an encoder with a limited attention span, in combination with the attention-based decoder (AD) and CTC. We note that while the encoder has a limited attention span, the attention-based decoder sees the entire encoder representation. The model employing AD could not function without chunking, whereas the CTC model processed the entire speech at once and still outperformed the AD model.

\subsection{Long-Form MT}
\label{sec:lf_mt}
The primary objective of long-form MT is to enhance textual coherence, as conventional MT systems assume sentence independence. Early work explored a concatenation of previous \cite{tiedemann-scherrer-2017-neural, donato-etal-2021-diverse} and future sentences \cite{agrawal-etal-2018-contextual}. These works showed that MT models benefit from the extra context and better handle the inter-sentential discourse phenomena. However, the benefits diminish if the context grows beyond a few sentences \cite{agrawal-etal-2018-contextual, Kim_Tran_Ney_2019, fernandes-etal-2021-measuring}. This can be attributed to the limitations of attention mechanisms, where an extensive volume of irrelevant information can lead to confusion. %

Other body of work tries to model very long sequences directly. \citet{dai-etal-2019-transformer} introduced a recurrence mechanism and improved positional encoding scheme in the Transformer. %
Later work proposed an explicit compressed memory realized by a few dense vectors \cite{feng-etal-2022-learn}.

\subsection{Long-Form Offline ST}
\label{sec:lf_st}
Unlike written input text in long-form MT, speech input in the ST task lacks explicit information about segmentation. Therefore, the research in the area of long-form offline speech translation concentrates on two separate issues: (1) improving \emph{segmentation} into sentences, and (2) enhancing robustness through the use of larger \emph{context}. 

In the traditional cascaded approach with separate speech recognition and machine translation models, the work focused on segmentation strategies for the ASR transcripts.\footnote{ASR transcripts are traditionally normalized, i.e., they consist of lowercase words without punctuation.} The methods are usually based on re-introducing punctuation to the transcript \cite{lu2010better, rangarajan-sridhar-etal-2013-segmentation, cho2015punctuation, cho2017nmt}. However, these approaches suffer from ASR error propagation and disregard the source audio's acoustic information. This was addressed by \citet{iranzo-sanchez-etal-2020-direct}, however, the approach still requires an intermediate ASR transcript that is unavailable in E2E models.

An alternative approach involves source-speech-based segmentation. The early work focused on VAD segmentation. This is usually sub-optimal as speakers place pauses inside sentences, not necessarily between them (e.g., hesitations before words with high information content, \citealp{goldman1958speech}). To this end, researchers tried considering not only the presence of speech but also its length \cite{potapczyk-przybysz-2020-srpols, inaguma-etal-2021-espnet, gaido2021beyond}. Later studies tried to avoid VAD and focused on more linguistically-motivated approaches, e.g., ASR CTC to predict voiced regions \citet{gallego-etal-2021-end} or directly modeling the sentence segmentation \cite{tsiamas2022shas, fukuda2022speech}. 

To address the problem of inadequate segmentation, \citet{gaido20_interspeech} showed that context-aware ST is less prone to segmentation errors. In an extensive study of context-aware ST, \citet{zhang2021beyond} observed that context improves quality, but this holds only for a limited number of utterances.

\subsection{Long-Form Simultaneous ST}
\label{sec:lf_sst}
Research focusing on direct long-form simultaneous speech translation remains relatively scarce. The closest works are in long-form simultaneous MT. \citet{schneider-waibel-2020-towards} proposed a streaming MT model capable of translating unsegmented text input. This model could be theoretically adapted for speech input. However, it was later shown that this model exhibits huge latency \cite{iranzo-sanchez-etal-2022-simultaneous}. Another work \cite{iranzo-sanchez-etal-2022-simultaneous} explored the extended context and confirmed the findings from long-form MT and offline ST, demonstrating that using the previous context significantly enhances performance. They also confirmed that a too-long context leads to decreased translation quality.

Finally, the only direct SST model that claims to work on a possibly unbounded input is \citet{ma2021streaming}. The model utilizes a Transformer encoder with a restriction on self-attention, allowing it to attend solely to a memory bank and a small segment. Unfortunately, based on the reported experiments, whether the model was specifically evaluated in the long-form setting remains unclear.

\subsection{Evaluation}
\label{sec:eval}
Evaluation of SST is a complex problem as we have to consider not only the translation quality but also the latency. Additionally, in the long-form regime, segmentation becomes another obstacle.

The most commonly used metric for translation quality in speech translation is BLEU \cite{papineni2002bleu,post-2018-call}. Other metrics such as chr{F}++ \cite{popovic-2017-chrf} and a neural-based metric COMET \cite{rei-etal-2020-unbabels} can be applied, too. %

The other important property of an SST system is latency. There are two main types of latencies: computation-unaware (CU) and computation-aware (CA) latency. The computation-unaware latency measures the delay in emitting a translation token relative to the source, regardless of the actual computation time. Hence, CU latency allows for a fair comparison regardless of the hardware infrastructure. However, CU latency cannot penalize the evaluated system for extensive computation; hence, CA latency can offer a more realistic assessment.

Measuring latency relative to the source or reference in SST is quite difficult because of the reordering present in translation. Historically, latency metrics were first developed for simultaneous machine translation (i.e., the source is text rather than speech). The most common are average lagging (AL; \citealp{ma-etal-2019-stacl}) and differentiable average lagging (DAL; \citealp{cherry2019thinking}). Broadly speaking, they measure ``how much of the source was read by the system to translate a word''. The latency unit is typically a word. The speech community quickly adopted these metrics. %
Unfortunately, these metrics assume a uniform distribution of words and uniform length of these words in the speech source. Alternatively, \citet{ansari-etal-2021-sltev} proposed to use a statistical word alignment of the candidate translation with the corresponding source transcript. This theoretically allows for more precise latency evaluation, but it is unclear how the alignment errors impact the reliability.

In the unsegmented long-form setting, additional issues arise. In a typical ``short-form'' segmented setup, the SST model does inference on a pre-segmented input. However, the candidate and reference segmentation into sentences might differ in the long-form unsegmented regime. Traditionally, this issue was addressed by re-segmenting the hypothesis based on the reference \cite{matusov-etal-2005-evaluating}. After the re-segmentation, a standard sentence-level evaluation of translation quality and latency is done. It should be noted that the commonly used latency metrics (AL, DAL) cannot be used in the long-form regime \cite{iranzo-sanchez-etal-2021-stream-level} without the re-segmentation. Yet, recent work observed that the re-segmentation introduces errors \cite{amrhein-haddow-2022-dont}. This poses a risk of incorrect translation and quality assessment and remains an open research question.

\vspace{-0.3em}
\section{Thesis Goals}
\label{sec:future_work}

The goal of our thesis is to achieve a ``true'' long-form simultaneous speech translation. This section outlines the steps we will take to accomplish this goal.

\subsection{Data and Evaluation}
In our future research, we will mainly use the setup similar to the IWSLT shared tasks \cite{ansari-etal-2020-findings, anastasopoulos-etal-2021-findings, anastasopoulos-etal-2022-findings}, i.e., mostly single speaker data. Identical to the IWSLT, we will treat the TED data as an in-domain setting. We will consider domains such as parliamentary speeches (e.g., Europarl-ST \citealp{iranzo2020europarl}) for the out-of-domain setting. As for the languages, we will include a diverse set of language pairs. A good inspiration might be again the IWSLT, i.e., English-to-\{German, Japanese, Chinese\}. Challenging will be the long-form setting, as to the best of our knowledge, none of the available data is strictly long-form. Our preliminary review found that the original TED talks can be reconstructed from the MuST-C \cite{mustc}  development and test set available for English-to-\{German, Japanese, Chinese\} language pairs.

As highlighted in the literature review in \cref{sec:eval}, evaluating the long-form SST remains an open problem. The quality and latency evaluation metrics currently used are designed for sentence-level evaluation. We must re-segment the long hypotheses into sentences based on their word alignment with provided references to use these metrics in the long-form regime. Unfortunately, the re-segmentation introduces errors, which poses a risk to the evaluation reliability. To tackle this, we will investigate alternative evaluation strategies. One potential approach for reducing the alignment error could be to move the alignment to the sentence level rather than the word level and allow an $m$-to-$n$ mapping between the reference and proposed sentences, similar to the Gale–Church alignment algorithm \cite{gale1994program}, with a reasonably small $m$ and $n$ (e.g., $0 \leq m,n \leq 2$). To verify the effectiveness of this method, we need to compare its correlation with human evaluations. 

\subsection{Quality-latency tradeoff in SST}
\label{goals:improving}
The first step of our research concentrates on enhancing the quality-latency tradeoff, mainly in the traditional ``short-form'' simultaneous speech translation. We hope the insights and improvements from the short-form regime will translate into the long-form regime.

In the research done so far, we already successfully reviewed the possibility of ``onlinizing'' state-of-the-art offline speech translation models in \citet{polak-etal-2022-cuni}. Our observations indicated that the attention-based encoder-decoder (AED) models tend to over-generate. This not only affects the resulting quality but also negatively impacts the AL latency evaluation reliability. Therefore, we proposed an improved version of the AL metric, which was later independently proposed under name length-adaptive average lagging (LAAL; \citealp{papi-etal-2022-generation}). To remedy the over-generation problem, we proposed an improved version of the beam search algorithm in \citet{polak23_interspeech}. While this led to significant improvements in the quality-latency tradeoff, the decoding still relied on label-synchronous decoding. In \citet{polak-etal-2023-towards}, we proposed a novel SST policy dubbed ``CTC policy'' that uses the output of an auxiliary CTC layer to guide the decoding. The proposed CTC policy led to even greater improvements in quality and reduced the real-time factor to 50~\%.

Thus far, our research has focused primarily on the AED architecture. Nonetheless, recent findings \cite{anastasopoulos-etal-2022-findings,agrawal-etal-2023-findings} suggest that other approaches, such as transducers \cite{graves2012sequence}, yield competitive results. Nevertheless, it remains unclear which approach is the most advantageous for SST. Our goal will be to compare these architectures for SST. We will put a particular emphasis on architectures with latent alignments (e.g., transducers). Generally, the latent alignment models make a strong monotonic assumption on the mapping between the source and the target, which might be problematic for the translation, typically involving word reordering. Therefore, we will assess the alignment quality and potential applications (such as segmentation). 

\subsection{Towards the Long-Form SST via On-the-Fly Segmentation}
In the second stage, we will concentrate on the long-form SST by utilizing on-the-fly segmentation and short-form models from the previous stage.

Drawing inspiration from offline long-form ST, which primarily emphasizes segmentation, we consider direct segmentation modeling the most promising approach \cite{tsiamas-etal-2022-pretrained, fukuda2022speech}. The limitation of these approaches is that they do not allow out-of-the-box simultaneous inference. However, we believe their adaptation to the simultaneous regime should be relatively straightforward (e.g., using a unidirectional encoder) and a custom decoding strategy. The main challenge here will be integrating this segmentation with existing models, especially considering the quality-latency tradeoff.

Our hopes go even further: Can we train a model to translate and predict the segmentation at the same time? The translation already contains punctuation marks (full stop, exclamation, and question marks), so if we knew the alignment between the translation and the source speech, we could use this information to segment the utterances directly. %
Therefore, we will experiment with various alignment approaches and asses their applicability to the segmentation. The results of our initial investigation on on-the-fly separation with CTC outputs are available in \citet{polak2023long}.

However, we see another valuable use of direct speech-to-translation alignments --- dataset creation. Today, ST datasets are created using the cascaded approach \cite{iranzo2020europarl, mustc, salesky2021multilingual}. The source transcript is first forced-aligned to the speech, then the transcript is word-aligned to the translations, and finally, these two alignments are used to segment the source speech into sentences based on the punctuation in the translation. In fact, this approach has a critical drawback: it virtually eliminates all data without a source transcript, preventing the research community from utilizing potentially valuable data sources. It is also worth noting that some languages do not have a writing system, which makes the direct speech-to-translation alignment even more attractive. Therefore, if the alignments show promising results, we will explore the feasibility of E2E speech-to-translation dataset creation.

An additional question is how to accommodate long context in the simultaneous regime. As pointed out in \cref{sec:lf_mt,sec:lf_st,sec:lf_sst}, the performance usually drops with a context longer than a few sentences. Some solutions have been suggested \cite{Kim_Tran_Ney_2019, feng-etal-2022-learn}, but it remains unclear how to adapt these approaches for SST with the specifics of SST in mind (e.g., computational constraints, speech input).

\subsection{True Long-Form SST}
The ultimate goal of our work is to achieve true long-form simultaneous speech translation. In other words, we aim to develop an architecture capable of processing a potentially infinite stream of speech input without any segmentation or special inference algorithm, translating the speech directly into the target language in real time. Admittedly, this is a very ambitious goal. However, there is plenty of evidence that it is feasible. For example, in long-form ASR,  related work has already observed that the RNN-T and CTC architectures are capable of long-form regime \cite{chiu2019comparison, narayanan2019recognizing, lu2021input, zhang2023google, rekesh2023fast}. Arguably, speech recognition is simpler than speech translation because it monotonically transcribes speech without reordering. However, the literature also shows that an architecture like RNN-T can be used in the ``short-form'' offline and simultaneous ST \cite{yan-etal-2023-espnet}. 

Therefore, based on the previous work in speech recognition and translation, we will propose a novel architecture that will allow simultaneous speech translation of a possibly infinite stream of speech. We will take inspiration from the existing architectures but revise them for the specific needs of simultaneous ST. This will require a particular focus on speech-to-translation alignment so that the source speech and target translation do not get out of sync. This architecture will also contain a ``forgetting'' mechanism that will allow the storage of essential bits of context while preventing memory issues. Finally, we will address the train-test mismatch because current hardware and training methods do not permit models to fit long inputs.

\section{Conclusion}

In conclusion, this thesis proposal presents an overview of the challenges involved in simultaneous speech translation (SST).
The literature review highlighted the limited research on long-form speech translation. Our research sets out three main goals with an emphasis on long-form speech translation.
These include improving the general quality-latency tradeoff in SST, exploring long-form SST through segmented inference, and ultimately achieving true long-form SST modeling.
We placed these goals in the context of related work and outlined a clear strategy for achieving them.

\section*{Acknowledgments}
Peter would like to thank his supervisor, Ond\v{r}ej Bojar, for his insight and guidance, as well as the anonymous reviewers for their valuable suggestions. This work has received support from GAUK project 244523 of Charles University and partial support from grant 19-26934X (NEUREM3) of the Czech Science Foundation.

\bibliography{anthology,custom}

\begin{thebibliography}{77}
\expandafter\ifx\csname natexlab\endcsname\relax\def\natexlab#1{#1}\fi

\bibitem[{Agarwal et~al.(2023)Agarwal, Agrawal, Anastasopoulos, Bentivogli,
  Bojar, Borg, Carpuat, Cattoni, Cettolo, Chen, Chen, Choukri, Chronopoulou,
  Currey, Declerck, Dong, Duh, Est{\`e}ve, Federico, Gahbiche, Haddow, Hsu,
  Mon~Htut, Inaguma, Javorsk{\'y}, Judge, Kano, Ko, Kumar, Li, Ma, Mathur,
  Matusov, McNamee, P.~McCrae, Murray, Nadejde, Nakamura, Negri, Nguyen,
  Niehues, Niu, Kr.~Ojha, E.~Ortega, Pal, Pino, van~der Plas, Pol{\'a}k,
  Rippeth, Salesky, Shi, Sperber, St{\"u}ker, Sudoh, Tang, Thompson, Tran,
  Turchi, Waibel, Wang, Watanabe, and Zevallos}]{agrawal-etal-2023-findings}
Milind Agarwal, Sweta Agrawal, Antonios Anastasopoulos, Luisa Bentivogli,
  Ond{\v{r}}ej Bojar, Claudia Borg, Marine Carpuat, Roldano Cattoni, Mauro
  Cettolo, Mingda Chen, William Chen, Khalid Choukri, Alexandra Chronopoulou,
  Anna Currey, Thierry Declerck, Qianqian Dong, Kevin Duh, Yannick Est{\`e}ve,
  Marcello Federico, Souhir Gahbiche, Barry Haddow, Benjamin Hsu, Phu Mon~Htut,
  Hirofumi Inaguma, D{\'a}vid Javorsk{\'y}, John Judge, Yasumasa Kano, Tom Ko,
  Rishu Kumar, Pengwei Li, Xutai Ma, Prashant Mathur, Evgeny Matusov, Paul
  McNamee, John P.~McCrae, Kenton Murray, Maria Nadejde, Satoshi Nakamura,
  Matteo Negri, Ha~Nguyen, Jan Niehues, Xing Niu, Atul Kr.~Ojha, John
  E.~Ortega, Proyag Pal, Juan Pino, Lonneke van~der Plas, Peter Pol{\'a}k,
  Elijah Rippeth, Elizabeth Salesky, Jiatong Shi, Matthias Sperber, Sebastian
  St{\"u}ker, Katsuhito Sudoh, Yun Tang, Brian Thompson, Kevin Tran, Marco
  Turchi, Alex Waibel, Mingxuan Wang, Shinji Watanabe, and Rodolfo Zevallos.
  2023.
\newblock \href {https://aclanthology.org/2023.iwslt-1.1} {{FINDINGS} {OF}
  {THE} {IWSLT} 2023 {EVALUATION} {CAMPAIGN}}.
\newblock In \emph{Proceedings of the 20th International Conference on Spoken
  Language Translation (IWSLT 2023)}, pages 1--61, Toronto, Canada (in-person
  and online). Association for Computational Linguistics.

\bibitem[{Agrawal et~al.(2018)Agrawal, Turchi, and
  Negri}]{agrawal-etal-2018-contextual}
Ruchit Agrawal, Marco Turchi, and Matteo Negri. 2018.
\newblock \href {https://aclanthology.org/2018.eamt-main.1} {Contextual
  handling in neural machine translation: Look behind, ahead and on both
  sides}.
\newblock In \emph{Proceedings of the 21st Annual Conference of the European
  Association for Machine Translation}, pages 31--40, Alicante, Spain.

\bibitem[{Amrhein and Haddow(2022)}]{amrhein-haddow-2022-dont}
Chantal Amrhein and Barry Haddow. 2022.
\newblock \href {https://aclanthology.org/2022.wmt-1.13} {Don{'}t discard
  fixed-window audio segmentation in speech-to-text translation}.
\newblock In \emph{Proceedings of the Seventh Conference on Machine Translation
  (WMT)}, pages 203--219, Abu Dhabi, United Arab Emirates (Hybrid). Association
  for Computational Linguistics.

\bibitem[{Anastasopoulos et~al.(2022)Anastasopoulos, Barrault, Bentivogli,
  Zanon~Boito, Bojar, Cattoni, Currey, Dinu, Duh, Elbayad, Emmanuel,
  Est{\`e}ve, Federico, Federmann, Gahbiche, Gong, Grundkiewicz, Haddow, Hsu,
  Javorsk{\'y}, Kloudov{\'a}, Lakew, Ma, Mathur, McNamee, Murray,
  N{\v{a}}dejde, Nakamura, Negri, Niehues, Niu, Ortega, Pino, Salesky, Shi,
  Sperber, St{\"u}ker, Sudoh, Turchi, Virkar, Waibel, Wang, and
  Watanabe}]{anastasopoulos-etal-2022-findings}
Antonios Anastasopoulos, Lo{\"\i}c Barrault, Luisa Bentivogli, Marcely
  Zanon~Boito, Ond{\v{r}}ej Bojar, Roldano Cattoni, Anna Currey, Georgiana
  Dinu, Kevin Duh, Maha Elbayad, Clara Emmanuel, Yannick Est{\`e}ve, Marcello
  Federico, Christian Federmann, Souhir Gahbiche, Hongyu Gong, Roman
  Grundkiewicz, Barry Haddow, Benjamin Hsu, D{\'a}vid Javorsk{\'y}, V{\u{e}}ra
  Kloudov{\'a}, Surafel Lakew, Xutai Ma, Prashant Mathur, Paul McNamee, Kenton
  Murray, Maria N{\v{a}}dejde, Satoshi Nakamura, Matteo Negri, Jan Niehues,
  Xing Niu, John Ortega, Juan Pino, Elizabeth Salesky, Jiatong Shi, Matthias
  Sperber, Sebastian St{\"u}ker, Katsuhito Sudoh, Marco Turchi, Yogesh Virkar,
  Alexander Waibel, Changhan Wang, and Shinji Watanabe. 2022.
\newblock \href {https://doi.org/10.18653/v1/2022.iwslt-1.10} {Findings of the
  {IWSLT} 2022 evaluation campaign}.
\newblock In \emph{Proceedings of the 19th International Conference on Spoken
  Language Translation (IWSLT 2022)}, pages 98--157, Dublin, Ireland (in-person
  and online). Association for Computational Linguistics.

\bibitem[{Anastasopoulos et~al.(2021)Anastasopoulos, Bojar, Bremerman, Cattoni,
  Elbayad, Federico, Ma, Nakamura, Negri, Niehues, Pino, Salesky, St{\"u}ker,
  Sudoh, Turchi, Waibel, Wang, and Wiesner}]{anastasopoulos-etal-2021-findings}
Antonios Anastasopoulos, Ond{\v{r}}ej Bojar, Jacob Bremerman, Roldano Cattoni,
  Maha Elbayad, Marcello Federico, Xutai Ma, Satoshi Nakamura, Matteo Negri,
  Jan Niehues, Juan Pino, Elizabeth Salesky, Sebastian St{\"u}ker, Katsuhito
  Sudoh, Marco Turchi, Alexander Waibel, Changhan Wang, and Matthew Wiesner.
  2021.
\newblock \href {https://doi.org/10.18653/v1/2021.iwslt-1.1} {{FINDINGS} {OF}
  {THE} {IWSLT} 2021 {EVALUATION} {CAMPAIGN}}.
\newblock In \emph{Proceedings of the 18th International Conference on Spoken
  Language Translation (IWSLT 2021)}, pages 1--29, Bangkok, Thailand (online).
  Association for Computational Linguistics.

\bibitem[{Ansari et~al.(2020)Ansari, Axelrod, Bach, Bojar, Cattoni, Dalvi,
  Durrani, Federico, Federmann, Gu, Huang, Knight, Ma, Nagesh, Negri, Niehues,
  Pino, Salesky, Shi, St{\"u}ker, Turchi, Waibel, and
  Wang}]{ansari-etal-2020-findings}
Ebrahim Ansari, Amittai Axelrod, Nguyen Bach, Ond{\v{r}}ej Bojar, Roldano
  Cattoni, Fahim Dalvi, Nadir Durrani, Marcello Federico, Christian Federmann,
  Jiatao Gu, Fei Huang, Kevin Knight, Xutai Ma, Ajay Nagesh, Matteo Negri, Jan
  Niehues, Juan Pino, Elizabeth Salesky, Xing Shi, Sebastian St{\"u}ker, Marco
  Turchi, Alexander Waibel, and Changhan Wang. 2020.
\newblock \href {https://doi.org/10.18653/v1/2020.iwslt-1.1} {{FINDINGS} {OF}
  {THE} {IWSLT} 2020 {EVALUATION} {CAMPAIGN}}.
\newblock In \emph{Proceedings of the 17th International Conference on Spoken
  Language Translation}, pages 1--34, Online. Association for Computational
  Linguistics.

\bibitem[{Ansari et~al.(2021)Ansari, Bojar, Haddow, and
  Mahmoudi}]{ansari-etal-2021-sltev}
Ebrahim Ansari, Ond{\v{r}}ej Bojar, Barry Haddow, and Mohammad Mahmoudi. 2021.
\newblock \href {https://doi.org/10.18653/v1/2021.eacl-demos.9} {{SLTEV}:
  Comprehensive evaluation of spoken language translation}.
\newblock In \emph{Proceedings of the 16th Conference of the European Chapter
  of the Association for Computational Linguistics: System Demonstrations},
  pages 71--79, Online. Association for Computational Linguistics.

\bibitem[{Arivazhagan et~al.(2020)Arivazhagan, Cherry, Macherey, and
  Foster}]{arivazhagan-etal-2020-translation}
Naveen Arivazhagan, Colin Cherry, Wolfgang Macherey, and George Foster. 2020.
\newblock \href {https://doi.org/10.18653/v1/2020.iwslt-1.27} {Re-translation
  versus streaming for simultaneous translation}.
\newblock In \emph{Proceedings of the 17th International Conference on Spoken
  Language Translation}, pages 220--227, Online. Association for Computational
  Linguistics.

\bibitem[{Bentivogli et~al.(2021)Bentivogli, Cettolo, Gaido, Karakanta,
  Martinelli, Negri, and Turchi}]{bentivogli-etal-2021-cascade}
Luisa Bentivogli, Mauro Cettolo, Marco Gaido, Alina Karakanta, Alberto
  Martinelli, Matteo Negri, and Marco Turchi. 2021.
\newblock \href {https://doi.org/10.18653/v1/2021.acl-long.224} {Cascade versus
  direct speech translation: Do the differences still make a difference?}
\newblock In \emph{Proceedings of the 59th Annual Meeting of the Association
  for Computational Linguistics and the 11th International Joint Conference on
  Natural Language Processing (Volume 1: Long Papers)}, pages 2873--2887,
  Online. Association for Computational Linguistics.

\bibitem[{Bojar et~al.(2021)Bojar, Mach{\'a}{\v{c}}ek, Sagar, Smr{\v{z}},
  Kratochv{\'\i}l, Pol{\'a}k, Ansari, Mahmoudi, Kumar, Franceschini, Canton,
  Simonini, Nguyen, Schneider, St{\"u}ker, Waibel, Haddow, Sennrich, and
  Williams}]{bojar-etal-2021-elitr}
Ond{\v{r}}ej Bojar, Dominik Mach{\'a}{\v{c}}ek, Sangeet Sagar, Otakar
  Smr{\v{z}}, Jon{\'a}{\v{s}} Kratochv{\'\i}l, Peter Pol{\'a}k, Ebrahim Ansari,
  Mohammad Mahmoudi, Rishu Kumar, Dario Franceschini, Chiara Canton, Ivan
  Simonini, Thai-Son Nguyen, Felix Schneider, Sebastian St{\"u}ker, Alex
  Waibel, Barry Haddow, Rico Sennrich, and Philip Williams. 2021.
\newblock \href {https://doi.org/10.18653/v1/2021.eacl-demos.32} {{ELITR}
  multilingual live subtitling: Demo and strategy}.
\newblock In \emph{Proceedings of the 16th Conference of the European Chapter
  of the Association for Computational Linguistics: System Demonstrations},
  pages 271--277, Online. Association for Computational Linguistics.

\bibitem[{Cattoni et~al.(2021)Cattoni, Di~Gangi, Bentivogli, Negri, and
  Turchi}]{mustc}
Roldano Cattoni, Mattia~Antonino Di~Gangi, Luisa Bentivogli, Matteo Negri, and
  Marco Turchi. 2021.
\newblock \href {https://doi.org/10.1016/j.csl.2020.101155} {Must-c: A
  multilingual corpus for end-to-end speech translation}.
\newblock \emph{Computer Speech \& Language}, 66:101155.

\bibitem[{Cherry and Foster(2019)}]{cherry2019thinking}
Colin Cherry and George Foster. 2019.
\newblock Thinking slow about latency evaluation for simultaneous machine
  translation.
\newblock \emph{arXiv preprint arXiv:1906.00048}.

\bibitem[{Chiu et~al.(2019)Chiu, Han, Zhang, Pang, Kishchenko, Nguyen,
  Narayanan, Liao, Zhang, Kannan et~al.}]{chiu2019comparison}
Chung-Cheng Chiu, Wei Han, Yu~Zhang, Ruoming Pang, Sergey Kishchenko, Patrick
  Nguyen, Arun Narayanan, Hank Liao, Shuyuan Zhang, Anjuli Kannan, et~al. 2019.
\newblock A comparison of end-to-end models for long-form speech recognition.
\newblock In \emph{2019 IEEE automatic speech recognition and understanding
  workshop (ASRU)}, pages 889--896. IEEE.

\bibitem[{Chiu et~al.(2021)Chiu, Narayanan, Han, Prabhavalkar, Zhang, Jaitly,
  Pang, Sainath, Nguyen, Cao et~al.}]{chiu2021rnn}
Chung-Cheng Chiu, Arun Narayanan, Wei Han, Rohit Prabhavalkar, Yu~Zhang,
  Navdeep Jaitly, Ruoming Pang, Tara~N Sainath, Patrick Nguyen, Liangliang Cao,
  et~al. 2021.
\newblock Rnn-t models fail to generalize to out-of-domain audio: Causes and
  solutions.
\newblock In \emph{2021 IEEE Spoken Language Technology Workshop (SLT)}, pages
  873--880. IEEE.

\bibitem[{Cho et~al.(2015)Cho, Niehues, Kilgour, and
  Waibel}]{cho2015punctuation}
Eunah Cho, Jan Niehues, Kevin Kilgour, and Alex Waibel. 2015.
\newblock Punctuation insertion for real-time spoken language translation.
\newblock In \emph{Proceedings of the 12th International Workshop on Spoken
  Language Translation: Papers}, pages 173--179.

\bibitem[{Cho et~al.(2017)Cho, Niehues, and Waibel}]{cho2017nmt}
Eunah Cho, Jan Niehues, and Alex Waibel. 2017.
\newblock Nmt-based segmentation and punctuation insertion for real-time spoken
  language translation.
\newblock In \emph{Interspeech}, pages 2645--2649.

\bibitem[{Cho and Esipova(2016)}]{cho2016can}
Kyunghyun Cho and Masha Esipova. 2016.
\newblock Can neural machine translation do simultaneous translation?
\newblock \emph{arXiv preprint arXiv:1606.02012}.

\bibitem[{Dai et~al.(2019)Dai, Yang, Yang, Carbonell, Le, and
  Salakhutdinov}]{dai-etal-2019-transformer}
Zihang Dai, Zhilin Yang, Yiming Yang, Jaime Carbonell, Quoc Le, and Ruslan
  Salakhutdinov. 2019.
\newblock \href {https://doi.org/10.18653/v1/P19-1285} {Transformer-{XL}:
  Attentive language models beyond a fixed-length context}.
\newblock In \emph{Proceedings of the 57th Annual Meeting of the Association
  for Computational Linguistics}, pages 2978--2988, Florence, Italy.
  Association for Computational Linguistics.

\bibitem[{Dalvi et~al.(2018)Dalvi, Durrani, Sajjad, and
  Vogel}]{dalvi-etal-2018-incremental}
Fahim Dalvi, Nadir Durrani, Hassan Sajjad, and Stephan Vogel. 2018.
\newblock \href {https://doi.org/10.18653/v1/N18-2079} {Incremental decoding
  and training methods for simultaneous translation in neural machine
  translation}.
\newblock In \emph{Proceedings of the 2018 Conference of the North {A}merican
  Chapter of the Association for Computational Linguistics: Human Language
  Technologies, Volume 2 (Short Papers)}, pages 493--499, New Orleans,
  Louisiana. Association for Computational Linguistics.

\bibitem[{Donato et~al.(2021)Donato, Yu, and Dyer}]{donato-etal-2021-diverse}
Domenic Donato, Lei Yu, and Chris Dyer. 2021.
\newblock \href {https://doi.org/10.18653/v1/2021.acl-long.104} {Diverse
  pretrained context encodings improve document translation}.
\newblock In \emph{Proceedings of the 59th Annual Meeting of the Association
  for Computational Linguistics and the 11th International Joint Conference on
  Natural Language Processing (Volume 1: Long Papers)}, pages 1299--1311,
  Online. Association for Computational Linguistics.

\bibitem[{Duong et~al.(2016)Duong, Anastasopoulos, Chiang, Bird, and
  Cohn}]{duong-etal-2016-attentional}
Long Duong, Antonios Anastasopoulos, David Chiang, Steven Bird, and Trevor
  Cohn. 2016.
\newblock \href {https://doi.org/10.18653/v1/N16-1109} {An attentional model
  for speech translation without transcription}.
\newblock In \emph{Proceedings of the 2016 Conference of the North {A}merican
  Chapter of the Association for Computational Linguistics: Human Language
  Technologies}, pages 949--959, San Diego, California. Association for
  Computational Linguistics.

\bibitem[{Feng et~al.(2022)Feng, Li, Song, Zheng, and
  Koehn}]{feng-etal-2022-learn}
Yukun Feng, Feng Li, Ziang Song, Boyuan Zheng, and Philipp Koehn. 2022.
\newblock \href {https://doi.org/10.18653/v1/2022.findings-naacl.105} {Learn to
  remember: Transformer with recurrent memory for document-level machine
  translation}.
\newblock In \emph{Findings of the Association for Computational Linguistics:
  NAACL 2022}, pages 1409--1420, Seattle, United States. Association for
  Computational Linguistics.

\bibitem[{Fernandes et~al.(2021)Fernandes, Yin, Neubig, and
  Martins}]{fernandes-etal-2021-measuring}
Patrick Fernandes, Kayo Yin, Graham Neubig, and Andr{\'e} F.~T. Martins. 2021.
\newblock \href {https://doi.org/10.18653/v1/2021.acl-long.505} {Measuring and
  increasing context usage in context-aware machine translation}.
\newblock In \emph{Proceedings of the 59th Annual Meeting of the Association
  for Computational Linguistics and the 11th International Joint Conference on
  Natural Language Processing (Volume 1: Long Papers)}, pages 6467--6478,
  Online. Association for Computational Linguistics.

\bibitem[{F{\"u}gen et~al.(2007)F{\"u}gen, Waibel, and
  Kolss}]{fugen2007simultaneous}
Christian F{\"u}gen, Alex Waibel, and Muntsin Kolss. 2007.
\newblock Simultaneous translation of lectures and speeches.
\newblock \emph{Machine translation}, 21:209--252.

\bibitem[{Fukuda et~al.(2022)Fukuda, Sudoh, and Nakamura}]{fukuda2022speech}
Ryo Fukuda, Katsuhito Sudoh, and Satoshi Nakamura. 2022.
\newblock Speech segmentation optimization using segmented bilingual speech
  corpus for end-to-end speech translation.
\newblock \emph{arXiv preprint arXiv:2203.15479}.

\bibitem[{Gaido et~al.(2020)Gaido, Gangi, Negri, Cettolo, and
  Turchi}]{gaido20_interspeech}
Marco Gaido, Mattia A.~Di Gangi, Matteo Negri, Mauro Cettolo, and Marco Turchi.
  2020.
\newblock \href {https://doi.org/10.21437/Interspeech.2020-2860}
  {{Contextualized Translation of Automatically Segmented Speech}}.
\newblock In \emph{Proc. Interspeech 2020}, pages 1471--1475.

\bibitem[{Gaido et~al.(2021)Gaido, Negri, Cettolo, and
  Turchi}]{gaido2021beyond}
Marco Gaido, Matteo Negri, Mauro Cettolo, and Marco Turchi. 2021.
\newblock Beyond voice activity detection: Hybrid audio segmentation for direct
  speech translation.
\newblock In \emph{Proceedings of the 4th International Conference on Natural
  Language and Speech Processing (ICNLSP 2021)}, pages 55--62.

\bibitem[{Gale et~al.(1994)Gale, Church et~al.}]{gale1994program}
William~A. Gale, Kenneth~Ward Church, et~al. 1994.
\newblock A program for aligning sentences in bilingual corpora.
\newblock \emph{Computational linguistics}, 19(1):75--102.

\bibitem[{G{\'a}llego et~al.(2021)G{\'a}llego, Tsiamas, Escolano, Fonollosa,
  and Costa-juss{\`a}}]{gallego-etal-2021-end}
Gerard~I. G{\'a}llego, Ioannis Tsiamas, Carlos Escolano, Jos{\'e} A.~R.
  Fonollosa, and Marta~R. Costa-juss{\`a}. 2021.
\newblock \href {https://doi.org/10.18653/v1/2021.iwslt-1.11} {End-to-end
  speech translation with pre-trained models and adapters: {UPC} at {IWSLT}
  2021}.
\newblock In \emph{Proceedings of the 18th International Conference on Spoken
  Language Translation (IWSLT 2021)}, pages 110--119, Bangkok, Thailand
  (online). Association for Computational Linguistics.

\bibitem[{Goldman-Eisler(1958)}]{goldman1958speech}
Frieda Goldman-Eisler. 1958.
\newblock Speech production and the predictability of words in context.
\newblock \emph{Quarterly Journal of Experimental Psychology}, 10(2):96--106.

\bibitem[{Graves(2012)}]{graves2012sequence}
Alex Graves. 2012.
\newblock Sequence transduction with recurrent neural networks.
\newblock \emph{arXiv preprint arXiv:1211.3711}.

\bibitem[{Graves et~al.(2006)Graves, Fern{\'a}ndez, Gomez, and
  Schmidhuber}]{graves2006connectionist}
Alex Graves, Santiago Fern{\'a}ndez, Faustino Gomez, and J{\"u}rgen
  Schmidhuber. 2006.
\newblock Connectionist temporal classification: labelling unsegmented sequence
  data with recurrent neural networks.
\newblock In \emph{Proceedings of the 23rd international conference on Machine
  learning}, pages 369--376.

\bibitem[{Harrison(2007)}]{harrison2007languages}
K~David Harrison. 2007.
\newblock \emph{When languages die: The extinction of the world's languages and
  the erosion of human knowledge}.
\newblock Oxford University Press.

\bibitem[{Huang et~al.(2022)Huang, Chang, Rybach, Prabhavalkar, Sainath,
  Allauzen, Peyser, and Lu}]{huang2022e2e}
W~Ronny Huang, Shuo-yiin Chang, David Rybach, Rohit Prabhavalkar, Tara~N
  Sainath, Cyril Allauzen, Cal Peyser, and Zhiyun Lu. 2022.
\newblock E2e segmenter: Joint segmenting and decoding for long-form asr.
\newblock \emph{arXiv preprint arXiv:2204.10749}.

\bibitem[{Inaguma et~al.(2021)Inaguma, Yan, Dalmia, Guo, Shi, Duh, and
  Watanabe}]{inaguma-etal-2021-espnet}
Hirofumi Inaguma, Brian Yan, Siddharth Dalmia, Pengcheng Guo, Jiatong Shi,
  Kevin Duh, and Shinji Watanabe. 2021.
\newblock \href {https://doi.org/10.18653/v1/2021.iwslt-1.10} {{ESP}net-{ST}
  {IWSLT} 2021 offline speech translation system}.
\newblock In \emph{Proceedings of the 18th International Conference on Spoken
  Language Translation (IWSLT 2021)}, pages 100--109, Bangkok, Thailand
  (online). Association for Computational Linguistics.

\bibitem[{Iranzo~Sanchez et~al.(2022)Iranzo~Sanchez, Civera, and
  Juan-C{\'\i}scar}]{iranzo-sanchez-etal-2022-simultaneous}
Javier Iranzo~Sanchez, Jorge Civera, and Alfons Juan-C{\'\i}scar. 2022.
\newblock \href {https://doi.org/10.18653/v1/2022.acl-long.480} {From
  simultaneous to streaming machine translation by leveraging streaming
  history}.
\newblock In \emph{Proceedings of the 60th Annual Meeting of the Association
  for Computational Linguistics (Volume 1: Long Papers)}, pages 6972--6985,
  Dublin, Ireland. Association for Computational Linguistics.

\bibitem[{Iranzo-S{\'a}nchez et~al.(2021)Iranzo-S{\'a}nchez, Civera~Saiz, and
  Juan}]{iranzo-sanchez-etal-2021-stream-level}
Javier Iranzo-S{\'a}nchez, Jorge Civera~Saiz, and Alfons Juan. 2021.
\newblock \href {https://doi.org/10.18653/v1/2021.findings-emnlp.58}
  {Stream-level latency evaluation for simultaneous machine translation}.
\newblock In \emph{Findings of the Association for Computational Linguistics:
  EMNLP 2021}, pages 664--670, Punta Cana, Dominican Republic. Association for
  Computational Linguistics.

\bibitem[{Iranzo-S{\'a}nchez et~al.(2020{\natexlab{a}})Iranzo-S{\'a}nchez,
  Gim{\'e}nez~Pastor, Silvestre-Cerd{\`a}, Baquero-Arnal, Civera~Saiz, and
  Juan}]{iranzo-sanchez-etal-2020-direct}
Javier Iranzo-S{\'a}nchez, Adri{\`a} Gim{\'e}nez~Pastor, Joan~Albert
  Silvestre-Cerd{\`a}, Pau Baquero-Arnal, Jorge Civera~Saiz, and Alfons Juan.
  2020{\natexlab{a}}.
\newblock \href {https://doi.org/10.18653/v1/2020.emnlp-main.206} {Direct
  segmentation models for streaming speech translation}.
\newblock In \emph{Proceedings of the 2020 Conference on Empirical Methods in
  Natural Language Processing (EMNLP)}, pages 2599--2611, Online. Association
  for Computational Linguistics.

\bibitem[{Iranzo-S{\'a}nchez et~al.(2020{\natexlab{b}})Iranzo-S{\'a}nchez,
  Silvestre-Cerda, Jorge, Rosell{\'o}, Gim{\'e}nez, Sanchis, Civera, and
  Juan}]{iranzo2020europarl}
Javier Iranzo-S{\'a}nchez, Joan~Albert Silvestre-Cerda, Javier Jorge, Nahuel
  Rosell{\'o}, Adria Gim{\'e}nez, Albert Sanchis, Jorge Civera, and Alfons
  Juan. 2020{\natexlab{b}}.
\newblock Europarl-st: A multilingual corpus for speech translation of
  parliamentary debates.
\newblock In \emph{ICASSP 2020-2020 IEEE International Conference on Acoustics,
  Speech and Signal Processing (ICASSP)}, pages 8229--8233. IEEE.

\bibitem[{Kim et~al.(2019)Kim, Tran, and Ney}]{Kim_Tran_Ney_2019}
Yunsu Kim, Duc~Thanh Tran, and Hermann Ney. 2019.
\newblock \href {https://doi.org/10.18653/v1/D19-6503} {When and why is
  document-level context useful in neural machine translation?}
\newblock In \emph{Proceedings of the Fourth Workshop on Discourse in Machine
  Translation (DiscoMT 2019)}, page 24–34, Hong Kong, China. Association for
  Computational Linguistics.

\bibitem[{Lu and Ng(2010)}]{lu2010better}
Wei Lu and Hwee~Tou Ng. 2010.
\newblock Better punctuation prediction with dynamic conditional random fields.
\newblock In \emph{Proceedings of the 2010 conference on empirical methods in
  natural language processing}, pages 177--186.

\bibitem[{Lu et~al.(2021)Lu, Pan, Doutre, Haghani, Cao, Prabhavalkar, Zhang,
  and Strohman}]{lu2021input}
Zhiyun Lu, Yanwei Pan, Thibault Doutre, Parisa Haghani, Liangliang Cao, Rohit
  Prabhavalkar, Chao Zhang, and Trevor Strohman. 2021.
\newblock Input length matters: Improving rnn-t and mwer training for long-form
  telephony speech recognition.
\newblock \emph{arXiv preprint arXiv:2110.03841}.

\bibitem[{Ma et~al.(2019)Ma, Huang, Xiong, Zheng, Liu, Zheng, Zhang, He, Liu,
  Li, Wu, and Wang}]{ma-etal-2019-stacl}
Mingbo Ma, Liang Huang, Hao Xiong, Renjie Zheng, Kaibo Liu, Baigong Zheng,
  Chuanqiang Zhang, Zhongjun He, Hairong Liu, Xing Li, Hua Wu, and Haifeng
  Wang. 2019.
\newblock \href {https://doi.org/10.18653/v1/P19-1289} {{STACL}: Simultaneous
  translation with implicit anticipation and controllable latency using
  prefix-to-prefix framework}.
\newblock In \emph{Proceedings of the 57th Annual Meeting of the Association
  for Computational Linguistics}, pages 3025--3036, Florence, Italy.
  Association for Computational Linguistics.

\bibitem[{Ma et~al.(2020)Ma, Dousti, Wang, Gu, and
  Pino}]{ma-etal-2020-simuleval}
Xutai Ma, Mohammad~Javad Dousti, Changhan Wang, Jiatao Gu, and Juan Pino. 2020.
\newblock \href {https://doi.org/10.18653/v1/2020.emnlp-demos.19} {{SIMULEVAL}:
  An evaluation toolkit for simultaneous translation}.
\newblock In \emph{Proceedings of the 2020 Conference on Empirical Methods in
  Natural Language Processing: System Demonstrations}, pages 144--150, Online.
  Association for Computational Linguistics.

\bibitem[{Ma et~al.(2021)Ma, Wang, Dousti, Koehn, and Pino}]{ma2021streaming}
Xutai Ma, Yongqiang Wang, Mohammad~Javad Dousti, Philipp Koehn, and Juan Pino.
  2021.
\newblock Streaming simultaneous speech translation with augmented memory
  transformer.
\newblock In \emph{ICASSP 2021-2021 IEEE International Conference on Acoustics,
  Speech and Signal Processing (ICASSP)}, pages 7523--7527. IEEE.

\bibitem[{Mach{\'a}{\v{c}}ek et~al.(2023)Mach{\'a}{\v{c}}ek, Dabre, and
  Bojar}]{machacek-etal-2023-turning}
Dominik Mach{\'a}{\v{c}}ek, Raj Dabre, and Ond{\v{r}}ej Bojar. 2023.
\newblock Turning whisper into real-time transcription system.
\newblock In \emph{Proceedings of the 3rd Conference of the Asia-Pacific
  Chapter of the Association for Computational Linguistics and the 13th
  International Joint Conference on Natural Language Processing: System
  Demonstrations}, Bali, Indonesia. Association for Computational Linguistics.

\bibitem[{Matusov et~al.(2005)Matusov, Leusch, Bender, and
  Ney}]{matusov-etal-2005-evaluating}
Evgeny Matusov, Gregor Leusch, Oliver Bender, and Hermann Ney. 2005.
\newblock \href {https://aclanthology.org/2005.iwslt-1.19} {Evaluating machine
  translation output with automatic sentence segmentation}.
\newblock In \emph{Proceedings of the Second International Workshop on Spoken
  Language Translation}, Pittsburgh, Pennsylvania, USA.

\bibitem[{Narayanan et~al.(2019)Narayanan, Prabhavalkar, Chiu, Rybach, Sainath,
  and Strohman}]{narayanan2019recognizing}
Arun Narayanan, Rohit Prabhavalkar, Chung-Cheng Chiu, David Rybach, Tara~N
  Sainath, and Trevor Strohman. 2019.
\newblock Recognizing long-form speech using streaming end-to-end models.
\newblock In \emph{2019 IEEE automatic speech recognition and understanding
  workshop (ASRU)}, pages 920--927. IEEE.

\bibitem[{Niehues et~al.(2016)Niehues, Nguyen, Cho, Ha, Kilgour, Müller,
  Sperber, Stüker, and Waibel}]{NiehuesNguyenCho2016_1000062876}
J.~Niehues, T.~S. Nguyen, E.~Cho, T.-L. Ha, K.~Kilgour, M.~Müller, M.~Sperber,
  S.~Stüker, and A.~Waibel. 2016.
\newblock \href {https://doi.org/10.21437/Interspeech.2016-154} {Dynamic
  transcription for low-latency speech translation}.
\newblock In \emph{17th Annual Conference of the International Speech
  Communication Association, INTERSPEECH 2016; Hyatt Regency San FranciscoSan
  Francisco; United States; 8 September 2016 through 16 September 2016}, volume
  08-12-September-2016 of \emph{Proceedings of the Annual Conference of the
  International Speech Communication Association. Ed. : N. Morgan}, pages
  2513--2517. {International Speech Communication Association}.

\bibitem[{Niehues et~al.(2018)Niehues, Pham, Ha, Sperber, and
  Waibel}]{NiehuesPhamHa2018_1000087584}
J.~Niehues, N.-Q. Pham, T.-L. Ha, M.~Sperber, and A.~Waibel. 2018.
\newblock \href {https://doi.org/10.21437/Interspeech.2018-1055} {Low-latency
  neural speech translation}.
\newblock In \emph{19th Annual Conference of the International Speech
  Communication, INTERSPEECH 2018; Hyderabad International Convention Centre
  (HICC)Hyderabad; India; 2 September 2018 through 6 September 2018. Ed.: C.C.
  Sekhar}, volume 2018-September of \emph{Proceedings of the Annual Conference
  of the International Speech Communication Association, INTERSPEECH}, pages
  1293--1297. {ISCA}.

\bibitem[{Osterholtz et~al.(1992)Osterholtz, Augustine, McNair, Rogina, Saito,
  Sloboda, Tebelskis, and Waibel}]{225935}
L.~Osterholtz, C.~Augustine, A.~McNair, I.~Rogina, H.~Saito, T.~Sloboda,
  J.~Tebelskis, and A.~Waibel. 1992.
\newblock \href {https://doi.org/10.1109/ICASSP.1992.225935} {Testing
  generality in janus: a multi-lingual speech translation system}.
\newblock In \emph{[Proceedings] ICASSP-92: 1992 IEEE International Conference
  on Acoustics, Speech, and Signal Processing}, volume~1, pages 209--212 vol.1.

\bibitem[{Papi et~al.(2022)Papi, Gaido, Negri, and
  Turchi}]{papi-etal-2022-generation}
Sara Papi, Marco Gaido, Matteo Negri, and Marco Turchi. 2022.
\newblock \href {https://doi.org/10.18653/v1/2022.autosimtrans-1.2}
  {Over-generation cannot be rewarded: Length-adaptive average lagging for
  simultaneous speech translation}.
\newblock In \emph{Proceedings of the Third Workshop on Automatic Simultaneous
  Translation}, pages 12--17, Online. Association for Computational
  Linguistics.

\bibitem[{Papineni et~al.(2002)Papineni, Roukos, Ward, and
  Zhu}]{papineni2002bleu}
Kishore Papineni, Salim Roukos, Todd Ward, and Wei-Jing Zhu. 2002.
\newblock Bleu: a method for automatic evaluation of machine translation.
\newblock In \emph{Proceedings of the 40th annual meeting of the Association
  for Computational Linguistics}, pages 311--318.

\bibitem[{Pol{\'a}k and Bojar(2023)}]{polak2023long}
Peter Pol{\'a}k and Ond{\v{r}}ej Bojar. 2023.
\newblock Long-form end-to-end speech translation via latent alignment
  segmentation.
\newblock \emph{arXiv preprint arXiv:2309.11384}.

\bibitem[{Pol\'ak et~al.(2023{\natexlab{a}})Pol\'ak, Liu, Pham, Niehues,
  Waibel, and Bojar}]{polak-etal-2023-towards}
Peter Pol\'ak, Danni Liu, Ngoc-Quan Pham, Jan Niehues, Alexander Waibel, and
  Ond{\v{r}}ej Bojar. 2023{\natexlab{a}}.
\newblock \href {https://aclanthology.org/2023.iwslt-1.37} {Towards efficient
  simultaneous speech translation: {CUNI}-{KIT} system for simultaneous track
  at {IWSLT} 2023}.
\newblock In \emph{Proceedings of the 20th International Conference on Spoken
  Language Translation (IWSLT 2023)}, pages 389--396, Toronto, Canada
  (in-person and online). Association for Computational Linguistics.

\bibitem[{Pol{\'a}k et~al.(2022)Pol{\'a}k, Pham, Nguyen, Liu, Mullov, Niehues,
  Bojar, and Waibel}]{polak-etal-2022-cuni}
Peter Pol{\'a}k, Ngoc-Quan Pham, Tuan~Nam Nguyen, Danni Liu, Carlos Mullov, Jan
  Niehues, Ond{\v{r}}ej Bojar, and Alexander Waibel. 2022.
\newblock \href {https://doi.org/10.18653/v1/2022.iwslt-1.24} {{CUNI}-{KIT}
  system for simultaneous speech translation task at {IWSLT} 2022}.
\newblock In \emph{Proceedings of the 19th International Conference on Spoken
  Language Translation (IWSLT 2022)}, pages 277--285, Dublin, Ireland
  (in-person and online). Association for Computational Linguistics.

\bibitem[{Pol\'ak et~al.(2023{\natexlab{b}})Pol\'ak, Yan, Watanabe, Waibel, and
  Bojar}]{polak23_interspeech}
Peter Pol\'ak, Brian Yan, Shinji Watanabe, Alexander Waibel, and Ondrej Bojar.
  2023{\natexlab{b}}.
\newblock {Incremental Blockwise Beam Search for Simultaneous Speech
  Translation with Controllable Quality-Latency Tradeoff}.
\newblock In \emph{Proc. Interspeech 2023}.

\bibitem[{Popovi{\'c}(2017)}]{popovic-2017-chrf}
Maja Popovi{\'c}. 2017.
\newblock \href {https://doi.org/10.18653/v1/W17-4770} {chr{F}++: words helping
  character n-grams}.
\newblock In \emph{Proceedings of the Second Conference on Machine
  Translation}, pages 612--618, Copenhagen, Denmark. Association for
  Computational Linguistics.

\bibitem[{Post(2018)}]{post-2018-call}
Matt Post. 2018.
\newblock \href {https://doi.org/10.18653/v1/W18-6319} {A call for clarity in
  reporting {BLEU} scores}.
\newblock In \emph{Proceedings of the Third Conference on Machine Translation:
  Research Papers}, pages 186--191, Brussels, Belgium. Association for
  Computational Linguistics.

\bibitem[{Potapczyk and Przybysz(2020)}]{potapczyk-przybysz-2020-srpols}
Tomasz Potapczyk and Pawel Przybysz. 2020.
\newblock \href {https://doi.org/10.18653/v1/2020.iwslt-1.9} {{SRPOL}{'}s
  system for the {IWSLT} 2020 end-to-end speech translation task}.
\newblock In \emph{Proceedings of the 17th International Conference on Spoken
  Language Translation}, pages 89--94, Online. Association for Computational
  Linguistics.

\bibitem[{Radford et~al.(2023)Radford, Kim, Xu, Brockman, McLeavey, and
  Sutskever}]{radford2023robust}
Alec Radford, Jong~Wook Kim, Tao Xu, Greg Brockman, Christine McLeavey, and
  Ilya Sutskever. 2023.
\newblock Robust speech recognition via large-scale weak supervision.
\newblock In \emph{International Conference on Machine Learning}, pages
  28492--28518. PMLR.

\bibitem[{Rangarajan~Sridhar et~al.(2013)Rangarajan~Sridhar, Chen, Bangalore,
  Ljolje, and Chengalvarayan}]{rangarajan-sridhar-etal-2013-segmentation}
Vivek~Kumar Rangarajan~Sridhar, John Chen, Srinivas Bangalore, Andrej Ljolje,
  and Rathinavelu Chengalvarayan. 2013.
\newblock \href {https://aclanthology.org/N13-1023} {Segmentation strategies
  for streaming speech translation}.
\newblock In \emph{Proceedings of the 2013 Conference of the North {A}merican
  Chapter of the Association for Computational Linguistics: Human Language
  Technologies}, pages 230--238, Atlanta, Georgia. Association for
  Computational Linguistics.

\bibitem[{Rei et~al.(2020)Rei, Stewart, Farinha, and
  Lavie}]{rei-etal-2020-unbabels}
Ricardo Rei, Craig Stewart, Ana~C Farinha, and Alon Lavie. 2020.
\newblock \href {https://aclanthology.org/2020.wmt-1.101} {Unbabel{'}s
  participation in the {WMT}20 metrics shared task}.
\newblock In \emph{Proceedings of the Fifth Conference on Machine Translation},
  pages 911--920, Online. Association for Computational Linguistics.

\bibitem[{Rekesh et~al.(2023)Rekesh, Kriman, Majumdar, Noroozi, Juang,
  Hrinchuk, Kumar, and Ginsburg}]{rekesh2023fast}
Dima Rekesh, Samuel Kriman, Somshubra Majumdar, Vahid Noroozi, He~Juang,
  Oleksii Hrinchuk, Ankur Kumar, and Boris Ginsburg. 2023.
\newblock Fast conformer with linearly scalable attention for efficient speech
  recognition.
\newblock \emph{arXiv preprint arXiv:2305.05084}.

\bibitem[{Ruiz and Federico(2014)}]{ruiz2014assessing}
Nicholas Ruiz and Marcello Federico. 2014.
\newblock Assessing the impact of speech recognition errors on machine
  translation quality.
\newblock In \emph{Proceedings of the 11th Conference of the Association for
  Machine Translation in the Americas: MT Researchers Track}, pages 261--274.

\bibitem[{Salesky et~al.(2021)Salesky, Wiesner, Bremerman, Cattoni, Negri,
  Turchi, Oard, and Post}]{salesky2021multilingual}
Elizabeth Salesky, Matthew Wiesner, Jacob Bremerman, Roldano Cattoni, Matteo
  Negri, Marco Turchi, Douglas~W Oard, and Matt Post. 2021.
\newblock The multilingual tedx corpus for speech recognition and translation.
\newblock \emph{arXiv preprint arXiv:2102.01757}.

\bibitem[{Schneider and Waibel(2020)}]{schneider-waibel-2020-towards}
Felix Schneider and Alexander Waibel. 2020.
\newblock \href {https://doi.org/10.18653/v1/2020.iwslt-1.28} {Towards stream
  translation: Adaptive computation time for simultaneous machine translation}.
\newblock In \emph{Proceedings of the 17th International Conference on Spoken
  Language Translation}, pages 228--236, Online. Association for Computational
  Linguistics.

\bibitem[{Sperber and Paulik(2020)}]{sperber-paulik-2020-speech}
Matthias Sperber and Matthias Paulik. 2020.
\newblock \href {https://doi.org/10.18653/v1/2020.acl-main.661} {Speech
  translation and the end-to-end promise: Taking stock of where we are}.
\newblock In \emph{Proceedings of the 58th Annual Meeting of the Association
  for Computational Linguistics}, pages 7409--7421, Online. Association for
  Computational Linguistics.

\bibitem[{Tan and Lim(2018)}]{tan2018artificial}
Kar-Han Tan and Boon~Pang Lim. 2018.
\newblock The artificial intelligence renaissance: deep learning and the road
  to human-level machine intelligence.
\newblock \emph{APSIPA Transactions on Signal and Information Processing},
  7:e6.

\bibitem[{Tiedemann and Scherrer(2017)}]{tiedemann-scherrer-2017-neural}
J{\"o}rg Tiedemann and Yves Scherrer. 2017.
\newblock \href {https://doi.org/10.18653/v1/W17-4811} {Neural machine
  translation with extended context}.
\newblock In \emph{Proceedings of the Third Workshop on Discourse in Machine
  Translation}, pages 82--92, Copenhagen, Denmark. Association for
  Computational Linguistics.

\bibitem[{Tsiamas et~al.(2022{\natexlab{a}})Tsiamas, G{\'a}llego, Escolano,
  Fonollosa, and Costa-juss{\`a}}]{tsiamas-etal-2022-pretrained}
Ioannis Tsiamas, Gerard~I. G{\'a}llego, Carlos Escolano, Jos{\'e} Fonollosa,
  and Marta~R. Costa-juss{\`a}. 2022{\natexlab{a}}.
\newblock \href {https://doi.org/10.18653/v1/2022.iwslt-1.23} {Pretrained
  speech encoders and efficient fine-tuning methods for speech translation:
  {UPC} at {IWSLT} 2022}.
\newblock In \emph{Proceedings of the 19th International Conference on Spoken
  Language Translation (IWSLT 2022)}, pages 265--276, Dublin, Ireland
  (in-person and online). Association for Computational Linguistics.

\bibitem[{Tsiamas et~al.(2022{\natexlab{b}})Tsiamas, G{\'a}llego, Fonollosa,
  and Costa-juss{\`a}}]{tsiamas2022shas}
Ioannis Tsiamas, Gerard~I G{\'a}llego, Jos{\'e}~AR Fonollosa, and Marta~R
  Costa-juss{\`a}. 2022{\natexlab{b}}.
\newblock Shas: Approaching optimal segmentation for end-to-end speech
  translation.
\newblock \emph{arXiv preprint arXiv:2202.04774}.

\bibitem[{Wang et~al.(2022)Wang, Guo, Li, Qiao, Wang, Li, Su, Chen, Zhang, Tao,
  Yang, and Qin}]{wang-etal-2022-hw-tscs}
Minghan Wang, Jiaxin Guo, Yinglu Li, Xiaosong Qiao, Yuxia Wang, Zongyao Li,
  Chang Su, Yimeng Chen, Min Zhang, Shimin Tao, Hao Yang, and Ying Qin. 2022.
\newblock \href {https://doi.org/10.18653/v1/2022.iwslt-1.21} {The
  {HW}-{TSC}{'}s simultaneous speech translation system for {IWSLT} 2022
  evaluation}.
\newblock In \emph{Proceedings of the 19th International Conference on Spoken
  Language Translation (IWSLT 2022)}, pages 247--254, Dublin, Ireland
  (in-person and online). Association for Computational Linguistics.

\bibitem[{Yan et~al.(2023)Yan, Shi, Tang, Inaguma, Peng, Dalmia, Polak,
  Fernandes, Berrebbi, Hayashi, Zhang, Ni, Hira, Maiti, Pino, and
  Watanabe}]{yan-etal-2023-espnet}
Brian Yan, Jiatong Shi, Yun Tang, Hirofumi Inaguma, Yifan Peng, Siddharth
  Dalmia, Peter Polak, Patrick Fernandes, Dan Berrebbi, Tomoki Hayashi, Xiaohui
  Zhang, Zhaoheng Ni, Moto Hira, Soumi Maiti, Juan Pino, and Shinji Watanabe.
  2023.
\newblock \href {https://aclanthology.org/2023.acl-demo.38} {{ESP}net-{ST}-v2:
  Multipurpose spoken language translation toolkit}.
\newblock In \emph{Proceedings of the 61st Annual Meeting of the Association
  for Computational Linguistics (Volume 3: System Demonstrations)}, pages
  400--411, Toronto, Canada. Association for Computational Linguistics.

\bibitem[{Yoshimura et~al.(2020)Yoshimura, Hayashi, Takeda, and
  Watanabe}]{yoshimura2020end}
Takenori Yoshimura, Tomoki Hayashi, Kazuya Takeda, and Shinji Watanabe. 2020.
\newblock End-to-end automatic speech recognition integrated with ctc-based
  voice activity detection.
\newblock In \emph{ICASSP 2020-2020 IEEE International Conference on Acoustics,
  Speech and Signal Processing (ICASSP)}, pages 6999--7003. IEEE.

\bibitem[{Zhang et~al.(2021)Zhang, Titov, Haddow, and
  Sennrich}]{zhang2021beyond}
Biao Zhang, Ivan Titov, Barry Haddow, and Rico Sennrich. 2021.
\newblock Beyond sentence-level end-to-end speech translation: Context helps.
\newblock In \emph{Proceedings of the 59th Annual Meeting of the Association
  for Computational Linguistics and the 11th International Joint Conference on
  Natural Language Processing (Volume 1: Long Papers)}, pages 2566--2578.

\bibitem[{Zhang et~al.(2023)Zhang, Han, Qin, Wang, Bapna, Chen, Chen, Li,
  Axelrod, Wang et~al.}]{zhang2023google}
Yu~Zhang, Wei Han, James Qin, Yongqiang Wang, Ankur Bapna, Zhehuai Chen, Nanxin
  Chen, Bo~Li, Vera Axelrod, Gary Wang, et~al. 2023.
\newblock Google usm: Scaling automatic speech recognition beyond 100
  languages.
\newblock \emph{arXiv preprint arXiv:2303.01037}.

\end{thebibliography}
\bibliographystyle{acl_natbib}

\appendix

\end{document}